\documentclass[conference]{IEEEtran}
\renewcommand\IEEEkeywordsname{Keywords}
\IEEEoverridecommandlockouts
\usepackage[nocompress]{cite}
\usepackage{balance} 
\usepackage{amsmath,amssymb,amsfonts}
\usepackage{algorithmic}
\usepackage{graphicx}
\usepackage{textcomp}
\usepackage{xcolor}
\usepackage{amsmath,amssymb,amsfonts}
\usepackage[ruled,linesnumbered,commentsnumbered]{algorithm2e}
\usepackage{multicol}
\usepackage{algorithmic}
\usepackage{graphicx}
\usepackage{tabularx}
\usepackage{textcomp}
\usepackage{xcolor}
\usepackage{float}
\usepackage[super]{nth}
\usepackage{subcaption}
\usepackage{booktabs}
\usepackage{multirow}
\usepackage{siunitx}
\usepackage{ragged2e}
\usepackage{threeparttable}
\usepackage{array}
\usepackage[super]{nth}
\usepackage{arydshln} 

\makeatletter

\begin{document}

\title{LuNet: A Deep Neural Network for Network Intrusion Detection}
\author{\IEEEauthorblockN{Peilun Wu\IEEEauthorrefmark{1} and Hui Guo\IEEEauthorrefmark{2}}
\IEEEauthorblockA{
\IEEEauthorrefmark{1}\IEEEauthorrefmark{2}The University of New South Wales, Sydney, Australia\\
Email: \{\IEEEauthorrefmark{1}z5100023,
\IEEEauthorrefmark{2}huig\}@cse.unsw.edu.au}}

\maketitle

\begin{abstract}
Network attack is a significant security issue for modern society. 
From small mobile devices to large cloud platforms, almost all computing products, used in our daily life, are networked and potentially under the threat of network intrusion. 
With the fast-growing network users, network intrusions become more and more frequent, volatile and advanced. 
Being able to capture intrusions in time for such a large scale network is critical and very challenging.
To this end, the machine learning (or AI) based network intrusion detection (NID), due to its intelligent capability, has drawn increasing attention in recent years. 
Compared to the traditional signature-based approaches, the AI-based solutions are more capable of detecting variants of advanced network attacks.
However, the high detection rate achieved by the existing designs is usually accompanied by a high rate of false alarms, which may significantly discount the overall effectiveness of the intrusion detection system. 
In this paper, we consider the existence of spatial and temporal features in the network traffic data and propose a hierarchical CNN+RNN neural network, LuNet. 
In LuNet, the convolutional neural network (CNN) and the recurrent neural network (RNN) learn input traffic data in sync with a gradually increasing granularity such that both spatial and temporal features of the data can be effectively extracted.
Our experiments on two network traffic datasets show that compared to the state-of-the-art network intrusion detection techniques, LuNet not only offers a high level of detection capability but also has a much low rate of false positive-alarm. 
\end{abstract}
\hfill \break

\section{Introduction}
Networked computing becomes indispensable to people's life. 
From daily communications to commercial transactions, from small businesses to large enterprises, all activities can be or will soon be done through networked services. 
Any vulnerabilities in the networked devices and computing platforms can expose the whole network under various attacks and may bring about disastrous consequences. 
Hence, effective network intrusion detection (NID) solutions are ultimately essential to the modern society. 

Initial NID designs are signature-based, where each type of attacks should be manually studied beforehand, and the detection is performed based on the attack's signatures. 
This kind of approaches are, however, not suitable to the fast-growing network and cannot cope with attacks of increasing volume, complexity and volatility. 

For the security of such a large scale and ever-expanding network, we need an intrusion detection system that is not only able to quickly and correctly identify known attacks but also adaptive and intelligent enough for the unknown and evolved attacks, which leads to AI-based solutions.
The artificial intelligence gained from machine learning enables the detection system to discover network attacks without much need for human intervention \cite{mukherjee1994network}. 

So far, investigations on the AI-based solutions are mainly based on two schemes: anomaly detection and misuse detection. 
The anomaly detection identifies an attack based on its anomalies deviated from the profile of normal traffic. 
Nevertheless, this scheme may have a high false positive rate if the normal traffic is not well profiled, and the profile used in the detection is not fully representative \cite{sommer2010outside}. 
Furthermore, to obtain a fully representative normal traffic profile for a dynamically evolving and expanding network is unlikely possible.

The misuse detection, on the other hand, focuses on the abnormal behaviour directly. The scheme can learn features of attacks based on a labelled training dataset, where both normal and attacked traffic data are marked.
Given sufficient labelled data, a misuse-detection design can effectively generate an abstract hypothesis of the boundary between normal and malicious traffic. 
The hypothesis can then be used to detect attacks for future unknown traffic. 
Therefore the misuse detection is more feasible and effective than the anomaly detection \cite{sommer2010outside,garcia2009anomaly} and has been adopted in real-world systems, such as Google, Qihoo 360 and Aliyun.

However, the existing misuse detection designs still present a high false positive rate, which significantly limits the in-time detection efficiency, incurs large manual scrutiny workload, and potentially degrades the network-wide security.
In this paper, we address this issue, and we aim for an improved misuse detection design. 
Our contributions are summarized as follows:
\begin{itemize}
\item We present a hierarchical deep neural network, LuNet, that is made of multiple levels of combined convolution and recurrent sub-nets; 
At each level, the input data are learned by both CNN and RNN nets; 
As the learning progresses from the first level to the last level, the learning granularity becomes increasingly detailed. 
With such an arrangement, the synergy of CNN and RNN can be effectively exploited for both spatial and temporal feature extractions.
\item We provide an in-depth analysis and discussion for the configuration of LuNet so that a high learning efficiency can be achieved.  
\end{itemize}

We test our design on two network intrusion datasets, NSL-KDD and UNSW-NB15, and we demonstrate that our design offers a higher detection capability (namely, better detection rate and validation accuracy) while maintaining a significantly lower false positive rate when compared to a set of state-of-the-art machine learning based designs. 

The remainder of this paper is organized as follows: Section \ref{Related Work} provides a brief background of machine learning for network intrusion detection.
The design of LuNet is detailed in Section \ref{LuNet}.
The threat model in the datasets used for LuNet design is presented in Section \ref{Threat Model}. 
The evaluation and comparison of LuNet with other state-of-the-art designs are given in Section \ref{Evaluation}.
The paper is concluded in Section \ref{Conclusion}.

\section{Background and Related Work}\label{Related Work}
Many algorithms have been developed for machine learning. 
They can be classified into classical machine learning approaches and deep learning approaches. 
Some of them, relevant to network intrusion detection are briefly discussed below.
\subsection{Classical Machine Learning Approaches}
Among many machine learning approaches \cite{buczak2015survey}, kernel machines and ensemble classifiers are two effective schemes that can be considered for NID. 
Support Vector Machine (SVM) that has long been used for task classification is a typical example of kernel machines, and Radial Basis Function (RBF) kernel (also called Gaussian kernel) is the most used kernel function \cite{ahmad2018performance}.
SVM makes the data that cannot be linearly separated in the original space, separable by projecting the data to a higher-dimension feature space. 
Ensemble classifiers,  such as Adaptive Boosting \cite{hu2013online}, Random Forest \cite{zhang2008random}, are often constructed with multiple weak classifiers to avoid overfitting during training so that a more robust classification function can be achieved. 

However, both kernel machines and ensemble classifiers are not scalable to a large data set. 
Their validation accuracy rarely scales with the size of training data \cite{bengio2006curse}. 
Given a high volume of training data available from the large scale network, using these traditional machine learning approaches to train the intrusion detection system is not efficient.  
Furthermore, the traditional approaches learn input data only based on a given set of features; they cannot generalize features from raw data, and their learning efficiency highly relies on the features specified. 

\subsection{Deep Learning Approaches}
Deep learning organizes ``learning algorithms" in layers in the form of ``artificial neural network" that can learn and make intelligent decisions on its own. 

Multi-Layer Perceptron (MLP) \cite{pal1992multilayer} is an early class of feedforward deep neural network that utilizes the backpropagation algorithm to minimize the error rate during training. 
MLP was initially used to solve complex approximate problems for speech recognition, image recognition, and machine translation.

The real flourish and practical breakthrough of deep learning stems from two popular deep learning algorithms: the convolutional neural network (CNN) and recurrent neural network (RNN).
CNN can automatically extract features of raw data and has gained great successes for image recognition \cite{lecun1998gradient,szegedy2015going,he2016deep}. 
However, the feature map generated by CNN often manifests the spatial relations of data. 
CNN does not work very well for the data of long-range dependency. 

RNN, on the other hand, has an ability to extract the temporal features from the input data.
Long short-term memory (LSTM) is a popular RNN \cite{hochreiter1997long}.
It keeps a trend of the long-term relationship in the sequential data while being able to drop out short-lived temporal noises. 

Recently, a hierarchical convolutional and recurrent neural network, HAST-IDS \cite{wang2017hast} has been used to learn spatial and temporal features from the network traffic data. 

The LuNet proposed in this paper is also a hierarchical network. 
Our design is similar to HAST-IDS in that both utilize CNN and RNN for spatial-temporal feature extraction.
However, there are some major differences:  
\begin{enumerate}
\item HAST-IDS stacks all RNN layers after CNN layers, while LuNet has a hierarchy of combined CNN and RNN layers. 
\item In HAST-IDS, because the CNN hierarchy is placed before RNN, the deep CNN may drop out the temporal information embedded in the raw input data, which makes RNN learning ineffective. LuNet, on the other hand, synchronizes both the CNN learning and the RNN learning into multiple steps with the learning granularity being gradually increased from coarse to fine; 
Therefore, both spatial and temporal features can be adequately captured. 
\item We apply batch normalization between CNN and RNN so that the learning efficiency and accuracy can be further improved. 
\end{enumerate}
The design of LuNet is presented in the next section.

\section{LuNet}\label{LuNet}
\subsection{Overview Structure}
\begin{figure}[t]
\centerline{\includegraphics[width=\linewidth]{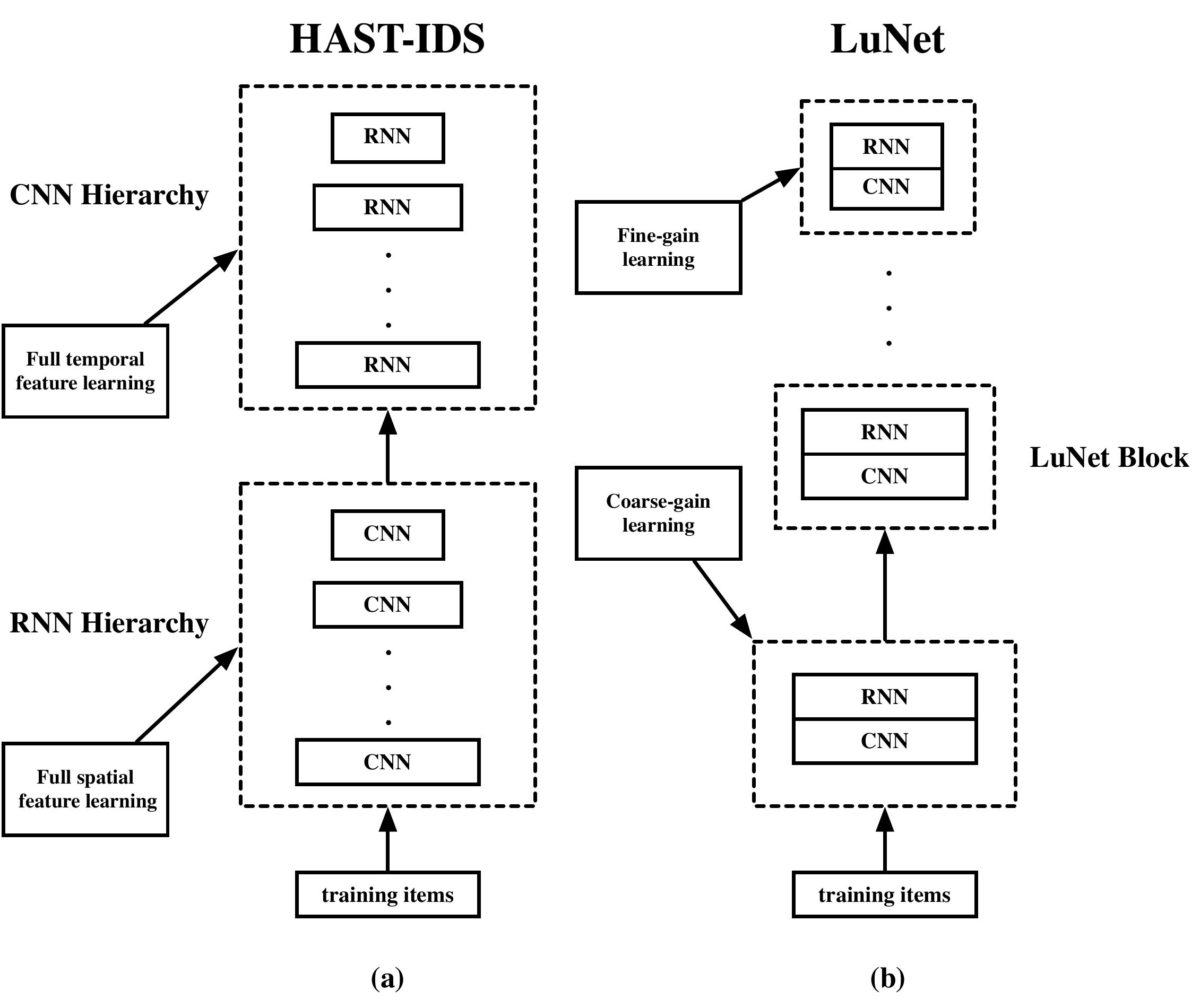}}
\caption{Different CNN and RNN arrangements. (a) HAST-IDS, CNN and RNN in tandem (b) LuNet, CNN and RNN mingled.}
\label{fig.1}
\end{figure}

As stated above, CNN targets on spatial features while RNN aims for temporal features.
The existing design HAST-IDS simply structures CNN and RNN in tandem, as illustrated in Fig.\ref{fig.1}(a). 
When the learning progresses along the multiple levels in the CNN hierarchy, the information extracted becomes more spatial oriented. The temporal features may be lost by the CNN hierarchy, which significantly limits the learning effectiveness of the following RNN (LSTM). 

Rather than allow CNN to learn to its full extent first,
in LuNet, we mingle the CNN and RNN subnets and synchronize both the CNN learning and the RNN learning into multiple steps and each step is performed by a combined CNN and RNN block, or \textit{LuNet block}, as illustrated in Fig.\ref{fig.1}(b). 
Since CNN is able to extract high-level features from a large amount of data, we place CNN before RNN at each level.
The learning starts from the first step on coarse-grain learning, hence the CNN output will still retain temporal information that will then be captured by RNN. 
The learning granularity becomes detailed as the data processing flows to the next step; but at each level, both CNN and RNN learn input on the same granularity. 
In such a way, CNN and RNN can learn to their full capacity without much interference with each other.

Given the overall structure of LuNet, the effectiveness of its learning is closely related to the hidden layers design, which is discussed in the next sub section.

\begin{figure}[t]
\centerline{\includegraphics[width=.85\linewidth]{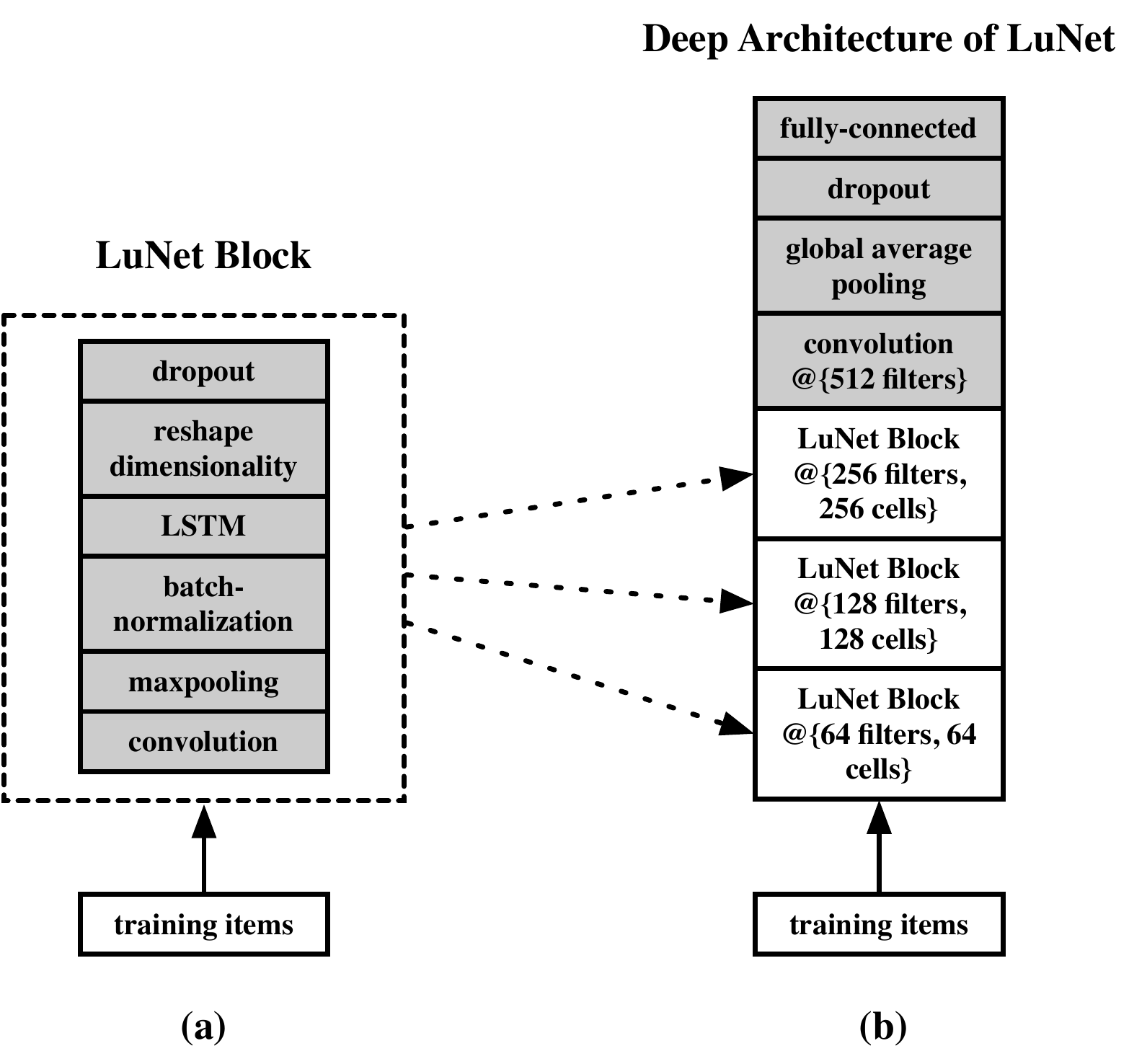}}
\caption{LuNet. (a) a LuNet Block. (b) the overall deep architecture of LuNet.}
\label{fig.LuNet}
\end{figure}
\subsection{Hidden Layers Design}

We measure the learning granularity in terms of the number of filters/cells used in the CNN/RNN network. 
The smaller the filter/cell number, the larger the granularity. 
Fig.\ref{fig.LuNet} shows a LuNet with three CNN+RNN levels.
The filter/cell number increases from 64 for the first level to 256 for the last level. 

The design considerations for each layer in a LuNet block and the final four processing layers for the learning outputs, as highlighted in shaded boxes in Fig.\ref{fig.LuNet}, are given below.


\subsubsection{Convolutional Neural Network (CNN)}

CNN mainly consists of two operations: convolution 
and pooling. 
Convolution transforms input data, through a set of filters or kernels, to an output that highlights the features of the input data, hence the output is usually called feature map.
The convolution output is further processed by an activation function 
and then down-sampled by pooling to trim off irrelevant data. 
Pooling also helps to remove glitches in the data to improve the learning for the following layers \cite{lecun2015deep,bengio2017deep}. 

CNN learns the input data by adjusting the filters automatically through rounds and rounds of learning processes so that its output feature map can effectively represent the raw input data. 

Since the network packet is presented in a 1D format, we use 1D convolution\footnote{One issue involved in convolution multiplication is to maintain the kernel size as same as the input, for which two typical schemes can be used: zero or no-zero padding. Since both schemes do not generate much difference to the learning outcome, here we choose no-zero padding for simplicity.} in LuNet, as illustrated 
in Formula~(\ref{formula:convolution}) that specifies the operation to the input vector $g$ with a  filter $f$ of size m. 
\begin{equation}
(f \ast g) (i) = \sum_{j=1}^{m}g(j) \cdot f(i-j+m/2), 
\label{formula:convolution}
\end{equation}
where $i$ is the position of different values in the sequence data.

Because the rectified linear unit (ReLU): $f(z) = max (0,z)$ is good for fast learning convergence, we therefore, choose it as the activation function. 
We also use the max pooling operation, as commonly applied in other existing designs \cite{zhou1988computation}.

\subsubsection{Batch Normalization}\label{Batch Normalization}

One problem with using the deep neural network is that the input value range dynamical changes from layer to layer during training, which is also known as covariance shift. The covariance shift causes the learning efficiency of one layer dependent on other layers, making the learning outcome unstable. Furthermore, because of the covariance shift, the learning rate is likely restricted to a low value to ensure data in different input ranges to be effectively learned, which slows down the learning speed. Batch normalization can be used to address this issue. 

Normalization scales data to the unit norm in the input layer and has been used to accelerate training deep neural network for image recognition \cite{ioffe2015batch}. 

We use batch normalization to adjust the CNN output for RNN in a LuNet block. The normalization subtracts the batch mean from each data and divides the result by the batch standard deviation, as given in Formula~(\ref{formula:normalization}).
\begin{equation}
\hat{x} = \frac{x-{\mu}_B}{\sqrt{{\delta}_B^2 +\epsilon}}, 
\label{formula:normalization}
\end{equation}
where $x$ is a value in the input batch, and ${\mu}_B$ and ${\delta}_B$ are, respectively, the batch mean and variance. The $\epsilon$ is an ignorable value, just to ensure the denominator in the formula non-zero.

Based on the normalized $\hat{x}$, the normalization produces the output y as given in Formula~(\ref{formula:normalization_output}), where the $\gamma$ and $\beta$ will be trained in the learning process for a better learning outcome.
\begin{equation}
\hat{y} = {\gamma}{\hat{x}}+{\beta}. 
\label{formula:normalization_output}
\end{equation}


\subsubsection{Long Short Term Memory (LSTM)}\label{LSTM}
Different from CNN that learns information on an individual-data-record basis, RNN can establish the relationship between data records by feeding back what has been learned from the previous learning to the current learning, and hence can capture the temporal features in the input data.

However, the simple feedback used in the traditional RNN may have a learning error accumulated in the long dependency. 
The accumulated errors may become large enough to invalid the final learning outcome. 
LSTM (Long Short-Term Memory), a gated recurrent neural network, mitigates such a problem. 
It controls the feedback with a set of gate functions such that the short-lived errors are eventually dropped out and only persistent features are retained. 
Therefore, we use LSTM for RNN.

\begin{figure}[t]
\centering
\centerline{\includegraphics[width=\linewidth]{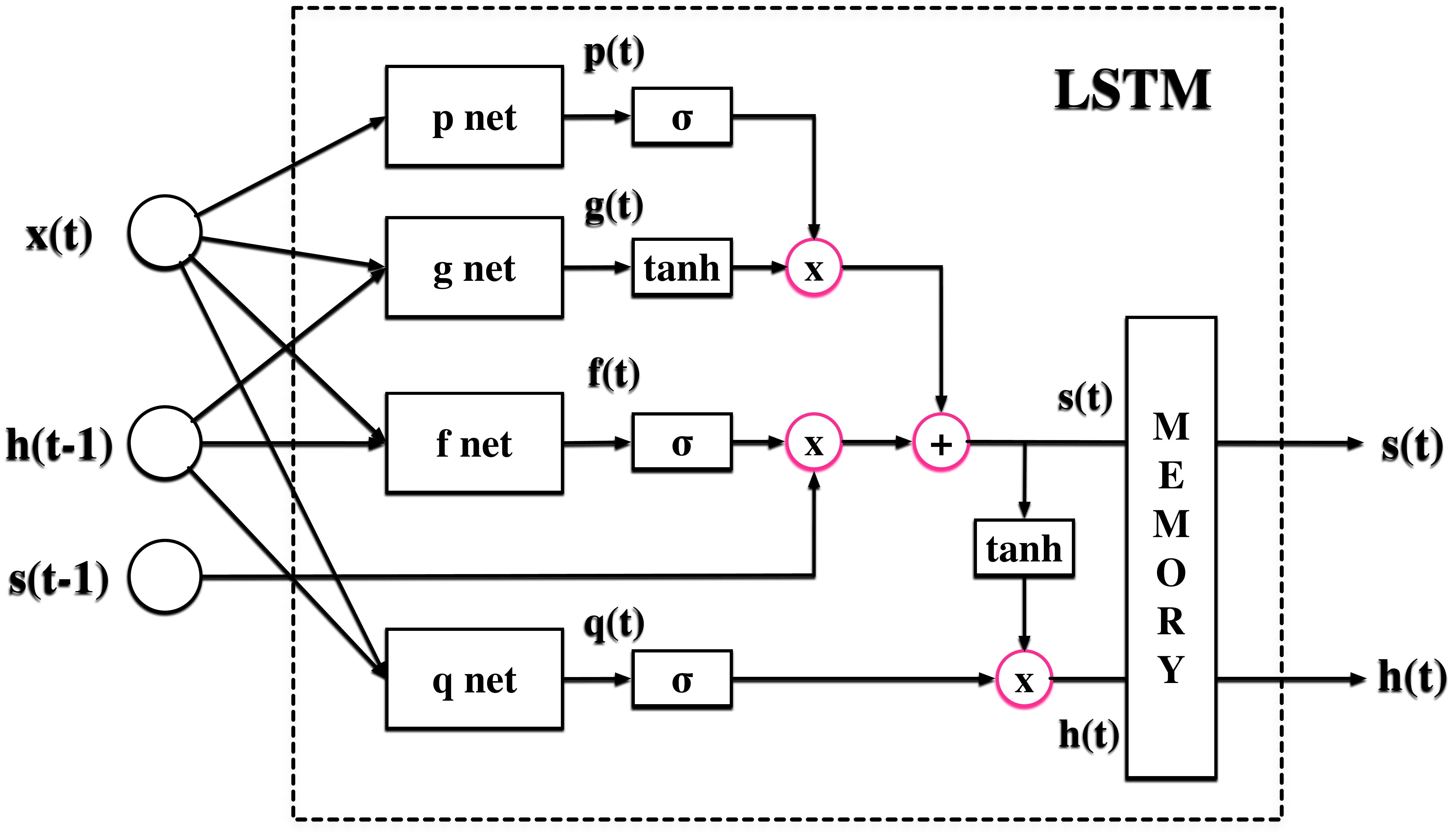}}
\caption{A high level data processing diagram of LSTM.}
\label{fig.lstm}
\end{figure}

For a brief description of LSTM operation, we create a high-level data processing diagram of LSTM \footnote{For the mathematical design details of LSTM, the reader is referred to the original paper \cite{bengio2017deep} as shown in Fig.~\ref{fig.lstm}.}

LSTM can be abstracted as a connection of four sub-networks (denoted as p-net, g-net, f-net and q-net in the diagram), a set of control gates, and a memory component. The input and output values in the diagram are vectors of the same size determined by the input $x(t)$. The state, $s(t)$, saved in the memory, serves as the feedback to the current learning. 

All the sub nets in LSTM have a similar structure, as specified in Formula~(\ref{formula.basic_LSTM_net}).
\begin{equation}
b+ U \times x(t)+ W \times h(t-1),
\label{formula.basic_LSTM_net}
\end{equation}
where $x(t)$, $h(t-1)$, $b$, $U$ and $W$ are, respectively, current input, previous output, bias, weight matrix for the current input, and recurrent weight matrix for the previous output. Each of the four nets have a different $b$, $U$, and $W$. 

The outputs from the sub nets ($p(t)$, $g(t)$, $f(t)$ and $q(t)$) are then used, through two types of controlling gates ($\sigma$ and $\tanh$) to determine the feedback s(t) from the previous learning and the current output h(t), as given in Formulas ~(\ref{formula.s(t)}) and (~\ref{formula.h(t)}), respectively.
\begin{equation}
s(t) = {\sigma}(f(t))*s(t-1) + {\sigma}(p(t))*\tanh{g(t)},
\label{formula.s(t)}
\end{equation}
\begin{equation}
h(t) = \tanh{s(t)}*{\sigma}(q(t)).
\label{formula.h(t)}
\end{equation}

LSTM learns the inputs by adjusting the weights in those nets and the $\sigma$ value such that the temporal features between the input data can be effectively generated in the output.

\subsubsection{Dimension Reshape}
Since in LuNet, the learning granularity changes from one CNN+RNN level to another, the output size of one level is different from the input size expected at the next level. 
We, therefore, add a layer to reshape the data for the next LuNet block. 

\subsubsection{Overfitting Prevention}
One typical problem when learning big data using a deep neural network is overfitting -- namely, the network has learned the training data too well, which restricts its ability to identify variants in new samples.
This problem can be handled by Dropout \cite{srivastava2014dropout}. 
Dropout randomly removes some connections from the deep neural network to reduce overfitting. 
We add the dropout layer with a default rate value of 0.5 
after the CNN+RNN hierarchy in LuNet.

\subsubsection{Final Layers}
Finally, an extra convolution layer and a global average pooling layer are used to extract further spatial-temporal features learned through the LuNet blocks.
The final learning output is generated by the last layer, a fully-connected layer.

\begin{figure}[t]
\begin{subfigure}{\linewidth}
\centerline{\includegraphics[width=\linewidth]{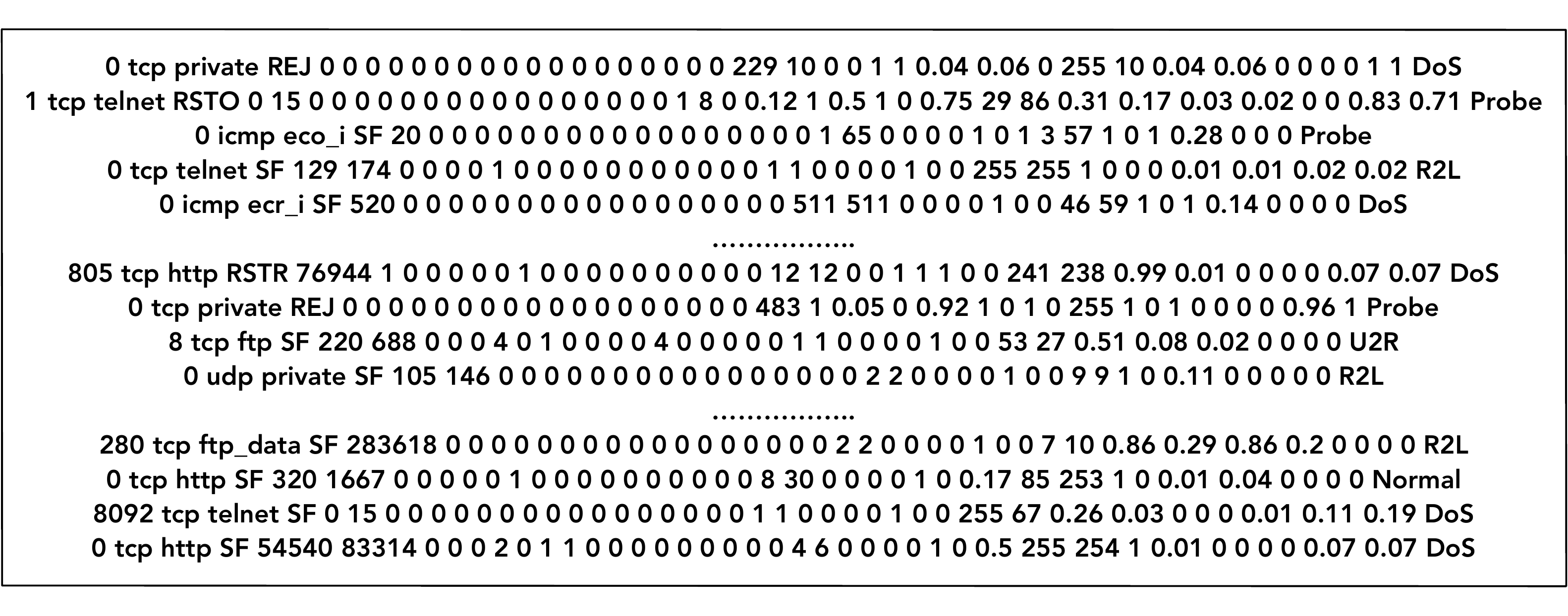}}
\caption{}
\end{subfigure}\\
\begin{subfigure}{\linewidth}
\centerline{\includegraphics[width=\linewidth]{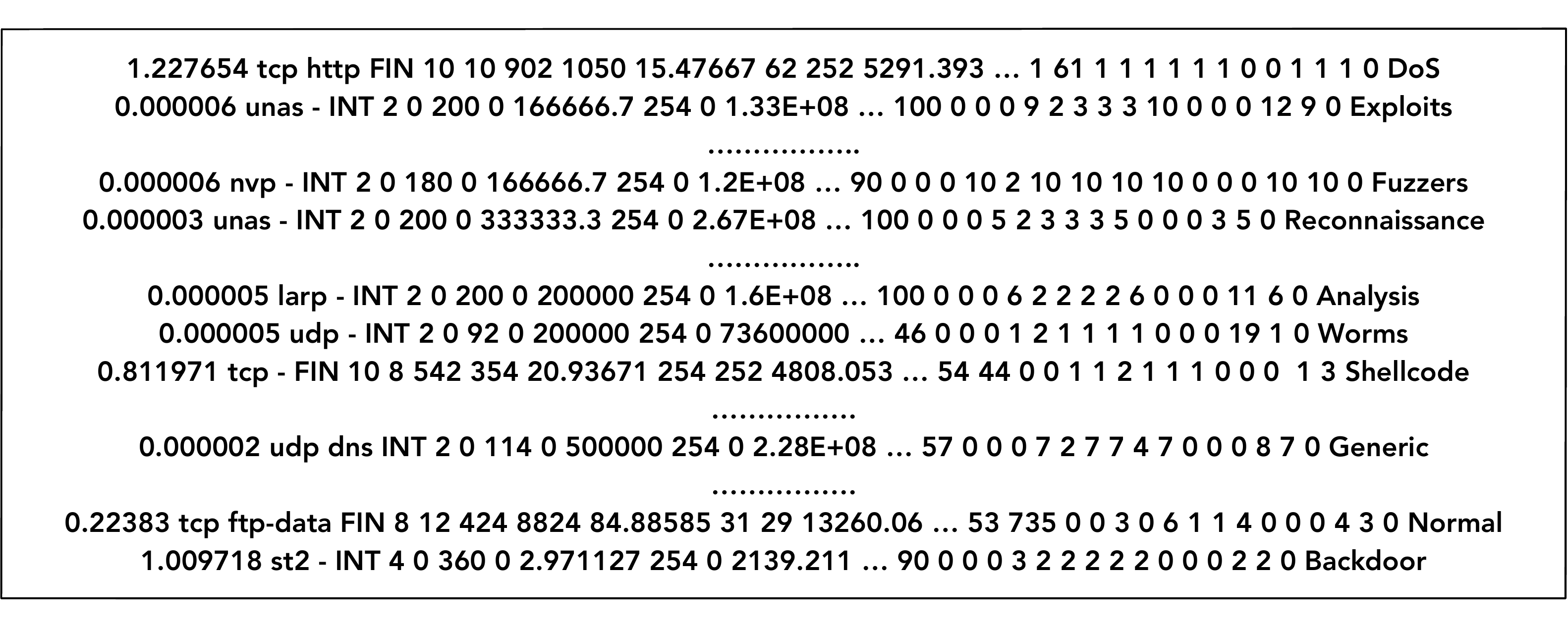}}
\caption{}
\end{subfigure}\\
\caption{Raw network traffic data. (a) NSL-KDD (41 features), (b) UNSW-NB15 (42 features).}
\label{fig.4}
\end{figure}

\section{Datasets and Threat Model}\label{Threat Model}

The evaluation of a neural network design is closely related to the dataset used.
Many datasets collected for NID contain significant amount of redundant data \cite{lippmann2000evaluating,mchugh2000testing}, which makes evaluation results unreliable \cite{ahmad2018performance,hu2013online,zhang2008random,wang2017hast,zhuo2017network}. 
To ensure the effectiveness of the evaluation, we select two non-redundant datasets: NSL-KDD and UNSW-NB15 in our investigation. 

NSL-KDD \cite{tavallaee2009detailed} was generated by \textit{Canadian Institute for Cyber Security (CICS)} from the original KDD'99 dataset. 
It consists of 39 different types of attacks.
The attacks are categorized into four groups: Denial of Service (DoS), User to Root (U2R), Remote to Local (R2L) and Probe.
A big issue with this dataset is its imbalanced distribution, which often leads to a high false positive rate due to insufficient data available for training\cite{li2017intrusion,lin2018character}. 
Here, we tackle this problem with a cross-validation scheme\footnote{This problem was also addressed in \cite{wu2019transfer} with a different focus and strategy.}.

UNSW-NB15\cite{moustafa2015unsw,moustafa2016evaluation}, generated by \textit{Australian Center for Cyber Security (ACCS)} in 2015, is a more contemporary dataset.
For the dataset, the attack samples were first collected from the three real-world websites: CVE (Common Vulnerabilities and Exposures)\footnote{CVE: https://cve.mitre.org/}, BID (Symantec Corporation)\footnote{BID: https://www.securityfocus.com}, and MSD (Microsoft Security Bulletin)\footnote{MSD: https://docs.microsoft.com/en-us/security-updates/securitybulletins}. The sample attacks were then simulated in a laboratory environment for the dataset generation.
There are nine attack categories in UNSW-NB15: DoS, Exploits, Generic, Shellcode, Reconnaissance, Backdoor, Worms, Analysis, and Fuzzers. 

\section{Evaluation and Discussion}\label{Evaluation}
To evaluate our design, we have implemented LuNet (as given in Fig.~\ref{fig.LuNet}) with TensorFlow backend, Keras and scikit-learn packages and we run the training on a HP EliteDesk 800 G2 SFF Desktop with Intel (R) Core (TM) i5-6500 CPU @ 3.20 GHz processor and 16.0 GB RAM. 
For comparison, we also implemented a set of state-of-the-art machine learning algorithms. 

The description of experiments and experiment results and discussion are presented below.


\subsection{Data Preprocessing}
\subsubsection{Convert Categorical Features}
For the experiment to be effective, we need data to conform to the input format required by the neural network.
Raw network traffic data include some categorical features as shown in Fig.\ref{fig.4}. The text information cannot be processed by a learning algorithm and should be converted into numerical values.
Here, we use the 'get\_dummies' function in Pandas \cite{mckinney-proc-scipy-2010} to operate the conversion.

\subsubsection{Standardization}
Input data may have varied distributions with different means and standard derivations, which may affect learning efficiency. We apply standardization to scale the input data to have a mean of 0 and a standard deviation of 1, as is often applied in many machine learning classifiers.


\subsubsection{Stratified K-Fold Cross Validation}
NSL-KDD and UNSW-NB15 contain 148,516 and 257,673 samples, respectively.
To realize the large non-redundant data for training and verification, we employ a Stratified K-Fold Cross Validation strategy, also commonly used in machine learning.
The scheme splits all samples in a dataset into $k$ groups; Among them, $k-1$ groups, as a whole, are used for training and the rest one group for validation; hence the strategy is also called \textit{Leave One Out Strategy}. 

\renewcommand{\baselinestretch}{1.5}
\begin{table}[t]
\centering
\begin{threeparttable}[htbp]
\centering
\caption{\textsc{Result of Binary Classification}}
 \begin{tabularx}{.92\linewidth}{c|ccc|ccc}
    \hline
    \multirow{2}{*}{$k$} &
      \multicolumn{3}{c|}{NSL-KDD} &
      \multicolumn{3}{c}{UNSW-NB15} \\
      & {DR\%} & {ACC\%} & {FPR\%} & {DR\%} & {ACC\%} & {FPR\%}\\
      \hline
    2 & 98.55 & 99.09 & 0.41 & 98.50 & 96.65 & 6.60\\
    4 & 98.56 & 99.11 & 0.37 & 98.04 & 97.51 & 3.44\\
    6 & 99.18 & 99.30 & 0.59 & 98.12 & 97.62 & 3.26\\
    8 & 99.27 & 99.34 & 0.59 & 97.78 & 97.67 & 2.54\\
    10 & 99.00 & 99.36 & 0.7 & 98.45 & 97.57 & 3.96\\
    \hline
    average & 99.42 & 99.24 & 0.53 & 98.18 & 97.40 & 3.96\\  
    \hline
  \end{tabularx}
\label{table3}
\end{threeparttable}
\end{table}
\renewcommand{\baselinestretch}{1.0}

\begin{figure}[t]
\centering
\centerline{\includegraphics[width=\linewidth]{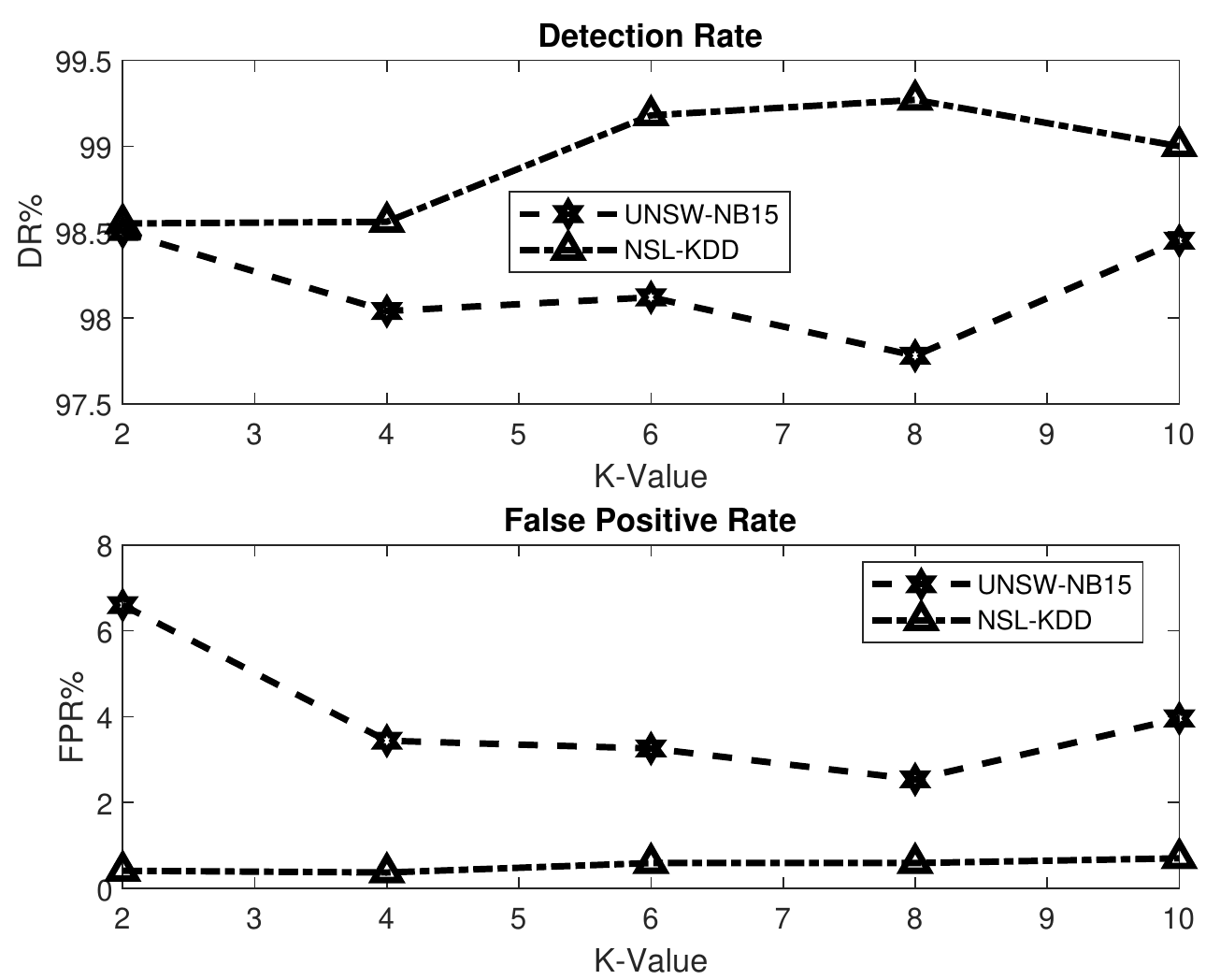}}
\caption{A comparison of detection rate (DR\%) and false positive rate (FPR\%) on using LuNet for binary classification based on NSL-KDD and UNSW-NB15 datasets.}
\label{fig.5}
\end{figure}

\begin{figure}[t]
\centering
\centerline{\includegraphics[width=\linewidth]{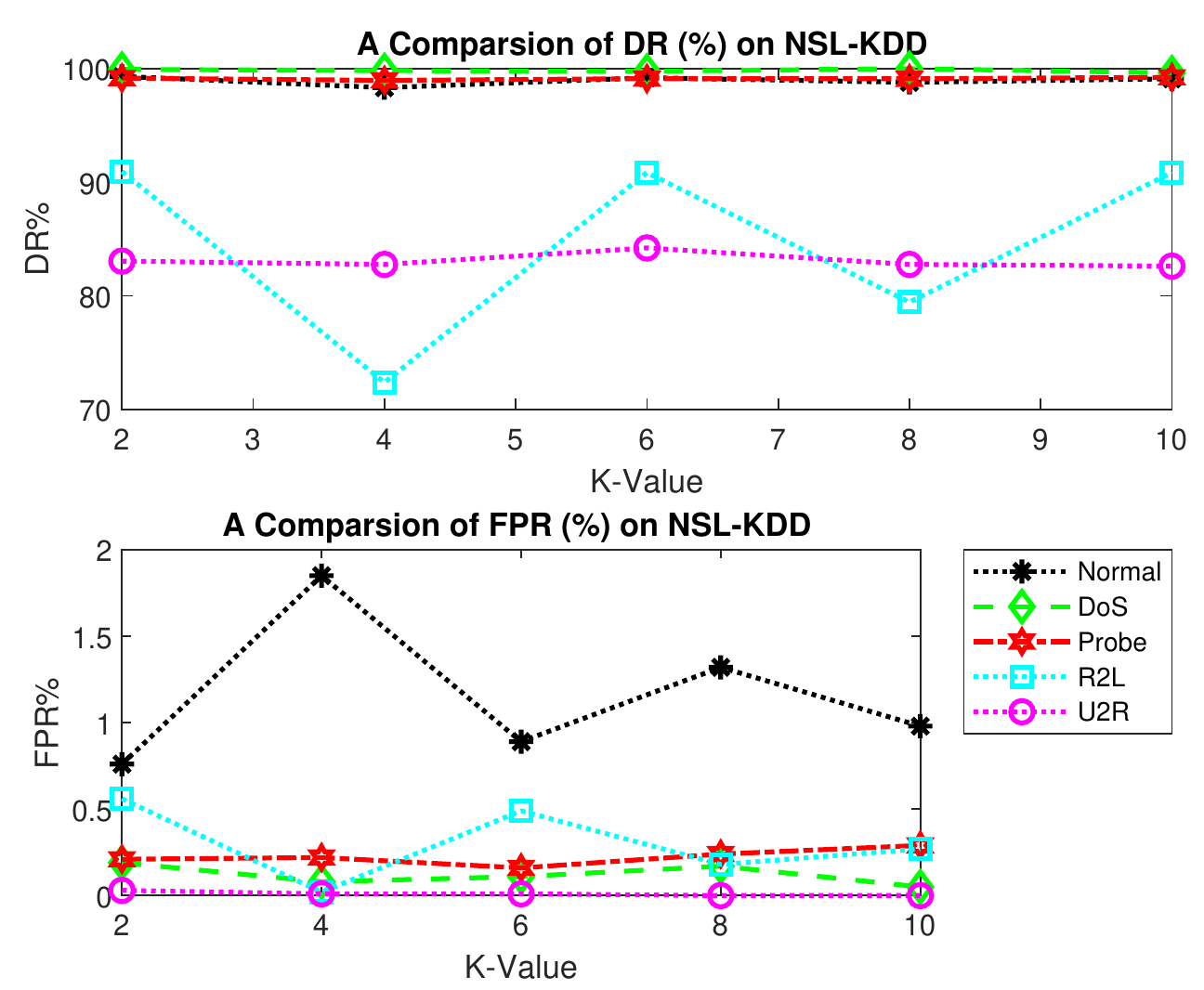}}
\caption{A comparison of detection rate (DR\%) and false positive rate (FPR\%) on using LuNet for multi-class classification based on NSL-KDD.}
\label{fig.6}
\end{figure}

\begin{figure}[t]
\centering
\centerline{\includegraphics[width=\linewidth]{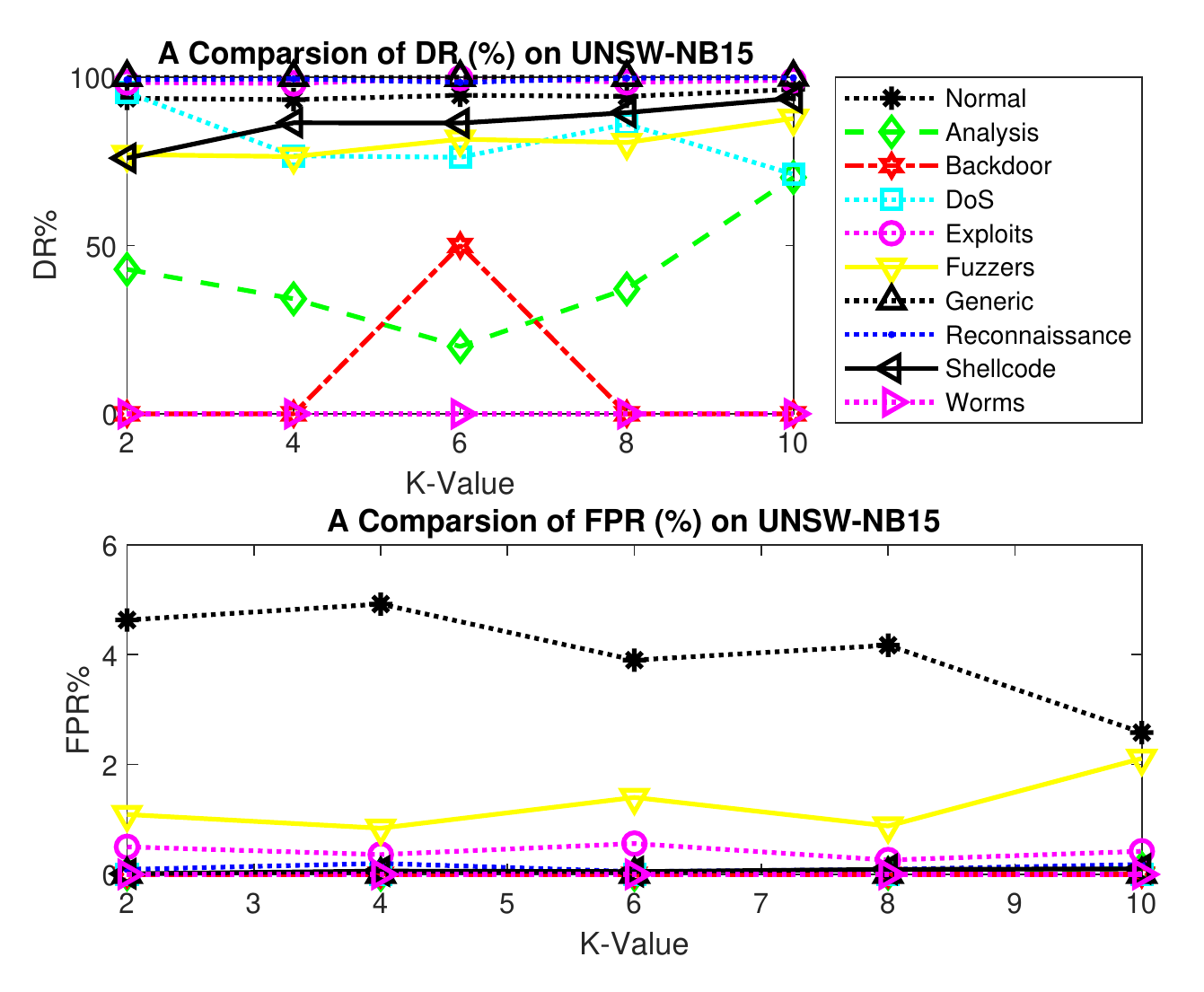}}
\caption{A comparison of detection rate (DR\%) and false positive rate (FPR\%) on using LuNet for multi-class classification based on UNSW-NB15.}
\label{fig.7}
\end{figure}

\subsection{Evaluation Metrics}
We evaluate LuNet in terms of the validation accuracy (ACC), detection rate (DR) and false positive rate (FPR). ACC measures LuNet's ability to correctly predict both attacked and non-attacked normal traffic, while DR indicates its ability of prediction for attacks only. A high DR can be shadowed by a high rate of false positive alarm (FPR), which therefore needs to be jointly considered with DR. The detail definitions of the three metrics are given in Formulas~(\ref{formula.ACC}), (\ref{formula.DR}) and (\ref{formula.FPR}).

\begin{equation}
{\small ACC = \dfrac{TP+TN}{TP+TN+FP+FN},}
\label{formula.ACC}
\end{equation}
\begin{equation}
{\small DR = \dfrac{TP}{TP+FN},}
\label{formula.DR}
\end{equation}
\begin{equation}
{\small FPR = \dfrac{FP}{FP+TN},}
\label{formula.FPR}
\end{equation}
where TP and TN are, respectively, the number of attacks and the number of normal traffic correctly classified; FP is the number of actual normal records mis-classified as attacks, and FN is the number of attacks falsely classified as normal traffic.

\subsection{LuNet Results}

In our experiments, we use RMSprop \cite{hinton2012neural} (a popular gradient descent algorithm) to optimize weights and biases for LuNet training, and for the learning rate, we set it to a median value 0.001 in a range given by the tensorflow library. We also choose the default rate value, 0.5, for the dropout layer. 

We first measure the performance of LuNet based on two scenarios: (1) binary classification, namely LuNet predicts a packet either as an attack or as a normal traffic; (2) multi-class classification, where LuNet identifies a packet either as normal or as one type of attacks given in the dataset attack model (namely, 5 classes for NSL-KDD and 10 classes for UNSW-NB15).
The experiment results are given below. 

\subsubsection{Binary Classification}
Table \ref{table3} shows the detection rate, accuracy and false positive rate for the binary classification of LuNet under different Stratified K-Fold Cross Validations, with k ranging from 2 to 10. 
The average values are given in the last row of the table.
As can be seen from the table, LuNet can achieve around 99.24\% and 97.40\% validation accuracy on NSL-KDD and UNSW-NB15, respectively.  
The best validation accuracy of 99.36\% on the NSL-KDD dataset and 97.67\% on the UNSW-NB15 can be observed.
It can also be seen that LuNet offers a high detection capability while incurs a low false positive alarm rate, as plotted Fig.\ref{fig.5}. 
On average, \textit{ DR=99.42\%} and \textit{FPR=0.53\%} for NSL-KDD, and \textit{DR=98.18\%} and \textit{FPR=3.96\%} on UNSW-NB15 can be obtained. 


It is worth to note that because of the cross-validation used, the false positive alarm rate is not affected by the imbalanced NSL-KDD dataset, as is expected.


\renewcommand{\baselinestretch}{1.5}
\begin{table}[t]
\centering
\begin{threeparttable}[htbp]
\centering
\caption{\textsc{Result of Multi-Class Classification}}
 \begin{tabularx}{.92\linewidth}{c|ccc|ccc}
    \hline
    \multirow{2}{*}{$k$} &
      \multicolumn{3}{c|}{NSL-KDD} &
      \multicolumn{3}{c}{UNSW-NB15} \\
      & {DR\%} & {ACC\%} & {FPR\%} & {DR\%} & {ACC\%} & {FPR\%}\\
      \hline
    2 & 99.24 & 99.09 & 0.98 & 95.37 & 84.61 & 1.69\\
    4 & 98.15 & 98.88 & 0.33 & 95.08 & 84.78 & 1.45\\
    6 & 99.11 & 99.13 & 0.77 & 96.10 & 84.93 & 2.09\\
    8 & 98.68 & 99.02 & 0.58 & 95.83 & 85.21 & 1.32\\
    10 & 99.02 & 99.14 & 0.61& 97.43 & 85.35 & 2.89\\ 
    \hline
    average & 98.84 & 99.05 & 0.65 & 95.96 & 84.98 & 1.89\\  
    \hline
  \end{tabularx}
\label{table4}
\end{threeparttable}
\end{table}
\renewcommand{\baselinestretch}{1.0}

\subsubsection{Multi-Class Classification}
Table \ref{table4} shows the multi-class classification results.
It can be seen from the table that LuNet can achieve an average of 99.05\% accuracy, 98.84\% detection rate on NSL-KDD, and 84.98\% accuracy and 95.96\% detection rate on UNSW-NB15 with, respectively, the 0.65\% and 1.89\% false positive rate on the two datasets.

Fig.\ref{fig.6} shows the detection rate (DR\%) and false positive rate (FPR\%) for 5 category (Normal, DoS, Probe, R2L, U2R) classification on NSL-KDD.
As we can see from the figure, LuNet shows an excellent capability to handle attacks in all categories (with a high detection rate and low false positive rate), except for U2R and R2L. 
For U2R and R2L, LuNet presents a relatively low detection rate, which indicates that the main features extracted by LuNet for U2R and R2L are not effectively distinct. 
There may be significant feature overlap between U2R and other attacks. 
The same reason is also applied to the Backdoor and Worms attacks, as manifested in the similar plot shown in  Fig.\ref{fig.7} for classification of 10 categories (Normal, Analysis, Backdoor, DoS, Exploits, Fuzzers, Generic, Reconnaissance, Shellcode, Worms) on UNSW-NB15.

\begin{figure}[t]
\centering
\centerline{\includegraphics[width=\linewidth]{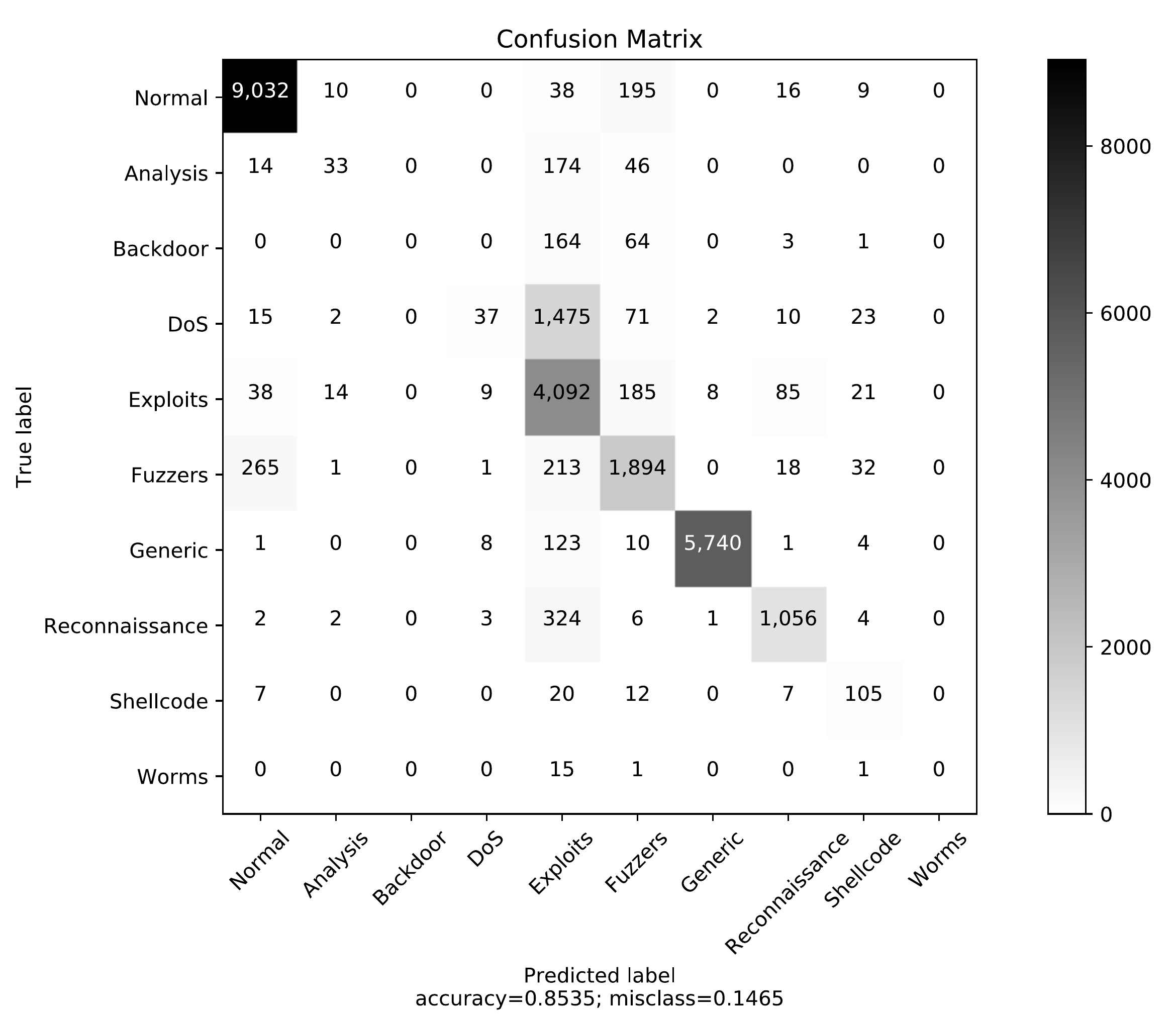}}
\caption{The confusion matrix of using LuNet on the UNSW-NB15 data set when $k=10$ for multi-class classification.}
\label{fig.8}
\end{figure}

\renewcommand{\baselinestretch}{1.5}
\begin{table}[t]
\centering
\caption{\linespread{1.0}\selectfont{} \textsc{A Comparison of 8 Algorithms for Multi-Class Classification Based on UNSW-NB15 Dataset with $K=10$ for Cross Validation}}
\begin{tabular}[\linewidth]{c|c||c|c|c}
\hline
\multicolumn{2}{c||}{Algorithm} & DR\%& ACC\%& FPR\%\\
\hline
\hline
\multirow{4}{*}{ML}& SVM (RBF) &83.71 & 74.80 & 7.73\\

\cdashline{2-5}
& RF &92.24 & 84.59 & 3.01\\
\cdashline{2-5}
& AdaBoost & 91.13 & 73.19 & 22.11\\
\cline{2-5}
& \textbf{average} &\textbf{89.03}&\textbf{77.53}&	\textbf{10.95}\\
\hline
\multirow{7}{*}{DL}& CNN &92.28 & 82.13 & 3.84\\
\cdashline{2-5}
& MLP & 96.74 &84.00 & 3.66\\
\cdashline{2-5}
& LSTM &92.76 & 82.40 & 3.63\\
\cdashline{2-5}
& HAST-IDS & 93.65 & 80.03 & 9.60\\
\cdashline{2-5}
& LuNet &97.43 & 85.35 & 2.89\\
\cline{2-5}
& \textbf{average} &\textbf{94.57}&	\textbf{82.78}&	\textbf{4.72}\\
\hline
\end{tabular}
\label{table5}
\end{table}
\renewcommand{\baselinestretch}{1.0}

Fig.\ref{fig.7} illustrates that LuNet can successfully detect most of the Normal traffic, Exploits, Generic, DoS, Shellcode and Reconnaissance attacks (see the overlapped lines at the very top of the plots) and has a moderate capability to detect Analysis attacks with the low false positive rate (FPR).
However, it is unable to discover Backdoor and Worms attacks, as can be observed in the confusion matrix (as shown in Fig.~\ref{fig.8}) generated from the experiment. We discover that most of Backdoor and Worms were, in fact, classified as Exploits attacks by LuNet.
The possible reason is that the Backdoor and Worms attacks use the common exploit flaws 
in the programs that listen for connections from remote hosts and the signature of those attacks is similar to the Exploits attacks. 
Therefore, the Backdoor and Worms attacks are treated as the  Exploits attacks by LuNet. 
Another possible reason may come from the insufficient data available for the Backdoor and Worms attacks; There are only around 1.4\% Backdoor attacks and 1.1\% Worms attacks in the dataset.


\subsection{Comparative Study}

We compare the performance of LuNet with other five state-of-the-art supervised learning algorithms that have been investigated 
in the literature: Support Vector Machine (SVM) with Gaussian Kernel (RBF)\cite{ahmad2018performance}, Multi-Layer Perceptron (MLP)\cite{esmaily2015intrusion}, Random Forest (RF)\cite{zhang2008random}, Adaptive Boosting (AdaBoost)\cite{hu2013online}, and HAST-IDS \cite{wang2017hast}.

We also run the experiment on individual CNN and LSTM networks, presented in LuNet.

The detection rate (DR\%), accuracy (ACC\%), and false positive rate (FPR\%) of the two design groups (ML, the classical machine learning algorithms and DL, the deep learning algorithms) are given in Table \ref{table5}. The average values of each group are also provided in the table. 

As can be seen from Table \ref{table5}, the deep neural networks generally have better performance than the three classical machine learning algorithms; On average, the deep neural networks have a higher detection rate (94.57\%) and accuracy (82.78\%) and a lower false positive rate (4.72\%), as compared to the corresponding average values of 89.03\%, 77.53\% and 10.95\% from the classical machine learning. Among the five deep learning networks, LuNet is better than other four DL algorithms due to the special combined CNN+RNN hierarchy used. In summary, LuNet demonstrates its superiority -- achieving a high detection rate and accuracy, while keeping a low false positive rate. 

\section{Conclusion}\label{Conclusion} 
In this paper, we present a deep neural network architecture, LuNet, to detect intrusions on a large scale network.
LuNet uses CNN to learn spatial features in the traffic data and LSTM for temporal features. 
To avoid the information loss due to different learning focuses of CNN and RNN, we synchronize both CNN and RNN to learn the input data at the same granularity. To enhance the learning, we also incorporate batch normalization in the design. 
Our experiments on the two non-redundant datasets, NSL-KDD and UNSW-NB15, show that LuNet is able to effectively take advantages of CNN and LSTM.
Compared with other state-of-the-art techniques, LuNet can significantly improve the validation accuracy and reduce the false positive rate for network intrusion detection.

It must be pointed out that LuNet does not work well to classify attacks of insufficient samples in the training dataset, such as Backdoors and Worms, as observed in our experiment,
which will be investigated in the future. 

\bibliographystyle{ieeetr}
\bibliography{ref.bib}

\end{document}